\documentclass[10pt,twocolumn,letterpaper]{article}
\usepackage{cvpr}
\usepackage{times}
\usepackage{epsfig}
\usepackage{graphicx}
\usepackage{amsmath}
\usepackage{amssymb}
\usepackage{array}
\usepackage{multirow}
\usepackage{color}
\newcommand{\modelname}{MA-RCA}
\newcommand{\norm}[1]{\left\lVert#1\right\rVert}
\DeclareMathOperator{\rank}{rank}
\DeclareMathOperator*{\argmin}{\arg\!\min}
\usepackage[pagebackref=true,breaklinks=true,letterpaper=true,colorlinks,bookmarks=false]{hyperref}
\usepackage{pdfcomment}
\usepackage[linesnumbered,ruled]{algorithm2e}

\newcommand{\mypdfc}[1]{}
\newcommand{\txg}[1]{}
\newcommand{\txr}[1]{}

\usepackage[pagebackref=true,breaklinks=true,letterpaper=true,colorlinks,bookmarks=false]{hyperref}

\cvprfinalcopy 

\def\cvprPaperID{3215} 

\begin{document}

\title{Multi-Attribute Robust Component Analysis for Facial UV Maps}

\author{Stylianos Moschoglou\textsuperscript{1,}\thanks{These two authors contributed equally.} , Evangelos Ververas\textsuperscript{1,}\footnotemark[1] , Yannis Panagakis\textsuperscript{1,2}, Mihalis Nicolaou\textsuperscript{3}, Stefanos Zafeiriou\textsuperscript{1}\\
\textsuperscript{1} Imperial College London, Department of Computing, London, UK\\
\textsuperscript{2} Middlesex University London, Department of Computing, London, UK\\ 
\textsuperscript{3} Goldsmiths, University of London, Department of Computing, London, UK\\
{\tt\small \{s.moschoglou, e.ververas16, i.panagakis, s.zafeiriou\}@imperial.ac.uk, m.nicolaou@gold.ac.uk}}

\maketitle

\begin{abstract}
Recently, due to the collection of large scale 3D face models, as well as the advent of deep learning, a significant progress has been made in the field of 3D face alignment ``in-the-wild''. That is, many methods have been proposed that establish sparse or dense 3D correspondences between a 2D facial image and a 3D face model. The utilization of 3D face alignment introduces new challenges and research directions, especially on the analysis of facial texture images. In particular, texture does not suffer any more from warping effects (that occurred when 2D face alignment methods were used). Nevertheless, since facial images are commonly captured in arbitrary recording conditions, a considerable amount of missing information and gross outliers is observed (e.g., due to self-occlusion, or subjects wearing eye-glasses). Given that many annotated databases have been developed for face analysis tasks, it is evident that component analysis techniques need to be developed in order to alleviate issues arising from the aforementioned challenges. In this paper, we propose a novel component analysis technique that is suitable for facial UV maps containing a considerable amount of missing information and outliers, while additionally, incorporates knowledge from various attributes (such as {\it age} and {\it identity}). We evaluate the proposed Multi-Attribute Robust Component Analysis (MA-RCA) on problems such as UV completion and age progression, where the proposed method outperforms compared techniques.  Finally, we demonstrate that MA-RCA method is powerful enough to provide weak annotations for training deep learning systems for various applications, such as illumination transfer.
\end{abstract}

\section{Introduction}
\begin{figure}[t]
\begin{center}
\includegraphics[width=1\linewidth]{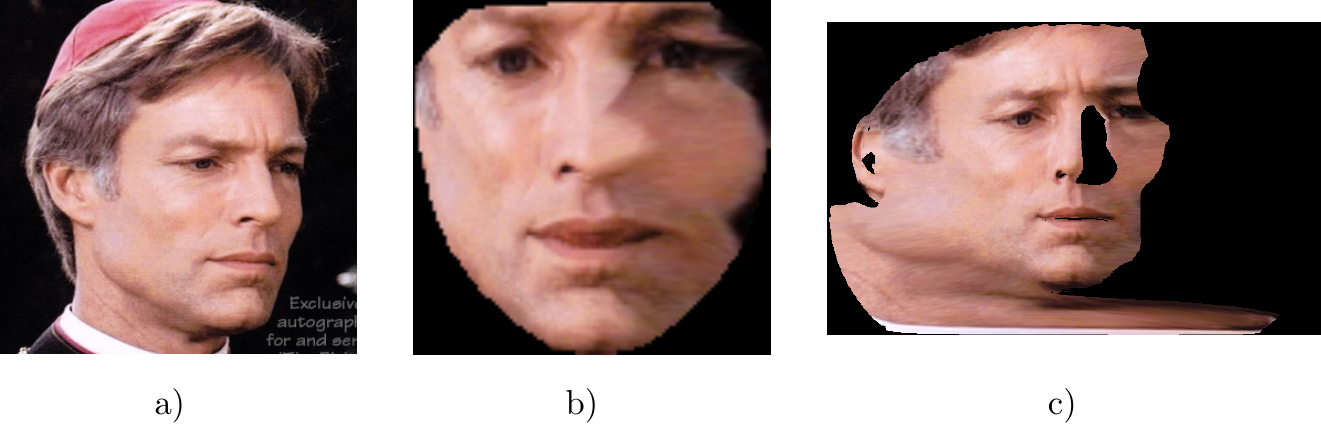}
\end{center}
   \caption{Original facial image is depicted in a). The facial image produced by a 2D face alignment algorithm is presented in b). The UV map derived by fitting a 3DMM is depicted in c). As can be clearly seen, 3DMM fitting overcomes drawbacks such as warping artifacts.}
\label{fig:1}
\end{figure}

Significant progress has been observed during the past years in the field of sparse and dense 3D face alignment \cite{jeni2016first,kim2017inversefacenet,zafeiriou20173d,zhu2016face,booth20173d}.  Recent developments include the utilization of Deep Neural Networks (DNNs) for estimation of 3D facial structure, as well as a methodology for fitting a 3D  Morphable Model (3DMM)  in ``in-the-wild'' images \cite{booth20173d}.  Additionally, several benchmarks for training sparse 3D face alignment models  have been recently developed \cite{zafeiriou20173d,booth20173d}.   The utilization of these methods introduces new challenges and opportunities as far as facial texture is concerned\txr{clarity}. In particular, by sampling over the fitted image, a 2D UV map of the facial texture can be constructed.   In order to further motivate the proposed method, in Fig. \ref{fig:1}, we depict an example of such a facial UV map.  As evinced by the figure, facial UV maps contain a considerable amount of missing data (pixels) due to factors such as self-occlusion.  Nevertheless, they do not suffer from warping effects, in contrast to the facial images produced by a 2D face alignment algorithm (Fig. \ref{fig:1}).  Utilizing facial UV maps for the discovery of latent components suitable for specific tasks (such as age or illumination transfer) requires the design of statistical component analysis methods that (a) can appropriately handle missing values, (b) can alleviate problems arising from gross errors, and (c) exploit any existing labels/attributes that are available. To tackle the aforementioned issues, it is natural to adopt techniques from the family of robust component analysis.

\begin{figure*}[t]
\begin{center}
\includegraphics[width=1\linewidth]{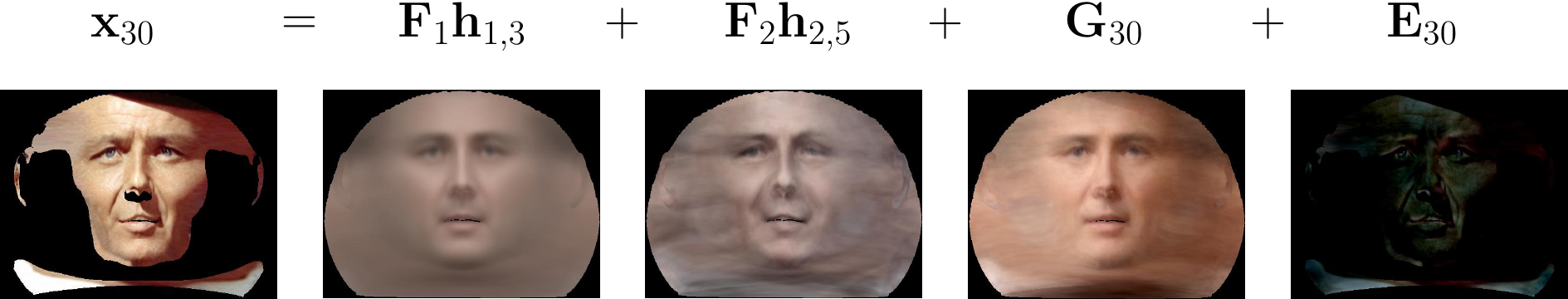}
\end{center}
   \caption{Facial UV map decomposition utilizing MA-RCA. Vectorized UV map $\mathbf{x}_{30}$ is decomposed into: a) an {\it age-group} component $\mathbf{F}_1\mathbf{h}_{1,3}$, where $\mathbf{F}_1$ is the base that renders the age-groups and $\mathbf{h}_{1,3}$ is the shared age selector corresponding to the third age-group in the training set (in this example, the age-group is {\it 41-50}), b) an {\it identity} component $\mathbf{F}_2\mathbf{h}_{2,5}$, where $\mathbf{F}_2$ is the base that renders the identities and $\mathbf{h}_{2,5}$ is the shared identity selector corresponding to the fifth identity in the training set (in this example, the identity is {\it Frank Sinatra}), c) an individual component $\mathbf{G}_{30}$ that captures all of the error-free information of the particular UV map that cannot be explained by the {\it age} and {\it identity} components, d) an error component $\mathbf{E}_{30}$ that captures the gross errors of the particular UV map (in this example, occlusions which is sparse, non-Gaussian noise).}
\label{fig:2}
\end{figure*}

\txr{we need to transition smoothly to related work}
In the past years, significant research has been conducted in terms of formulating robust component analysis techniques. Arguably, the most prominent example lies in the Robust PCA (RPCA) algorithm \cite{candes2011robust}, that has also been extended for handling missing values in \cite{wright2009robust}. The RPCA algorithm with missing values has been recently proven extremely useful towards the extraction of a low-rank sub-space of facial UV textures that is free of gross errors, thus deeming it extremely useful for the fitting of 3DMMs ``in-the-wild'' \cite{booth20173d}.  Nevertheless,  RPCA is an unsupervised component analysis technique, and hence does not take into account the various attributes/annotations that may be present in the data at-hand.

Other recent robust component analysis methods include Robust Correlated and Individual Component Analysis (RCICA) \cite{panagakis2016robust}, as well as the Robust Joint and Individual Variance Explained (RJIVE) \cite{sagonas2017robust}.  RCICA robustly recovers both the correlated and individual low-rank components of two views of noisy data, and can therefore be interpreted as a robust extension of Canonical Correlation Analysis (CCA) \cite{thompson2005canonical}.  Nevertheless, RCICA is not designed to utilize labels or any available annotations.  RJIVE further extends RCICA by extracting low-rank sub-spaces from multiple-views similarly to RCICA in the presence of a {\it single} attribute only (e.g., if the data at-hand are annotated for the attribute {\it age}, then they may be split in different age-groups and each age-group can be considered as a different view).  As a result, data that is annotated in terms of multiple attributes (such as {\it identity} and {\it age}) cannot be fully exploited in RJIVE.  Finally, both aforementioned methods have not been extended in order to deal with missing values, and thus can not be directly applied to the analysis of facial UV maps.

To alleviate the shortcomings of the previously mentioned methods with respect to the problem of facial UV analysis, in this paper we introduce Multi-Attribute Robust Component Analysis, dubbed MA-RCA. In summary, the contributions of the paper are as follows.
\begin{itemize}
 \item We introduce MA-RCA, a novel component analysis technique which recovers suitable components that robustly capture the shared and individual variation of data under a multi-attribute scenario. Furthermore, MA-RCA is inherently able to handle observations with missing values, as well as sparse and gross corruptions. 
 \item We demonstrate that MA-RCA can be applied to a number of challenging problems, such as  completion of missing data in the texture of a reconstructed 3D facial image and {\it transfer} of multiple attributes in images captured ``in-the-wild'' (e.g., {\it illumination}, {\it identity} and {\it age}). 
 \item We show that the components obtained by MA-RCA can be used to train deep learning systems for various tasks. 
 
\end{itemize}

The rest of the paper is organized as follows. In Section \ref{section_2} we provide the mathematical formulation of \modelname\ and present all the necessary optimization algorithms. In Section \ref{experiments} we run a series of experiments and demonstrate the merits of \modelname\ against other state-of-the-art algorithms.
\section{Multi-Attribute Robust Component Analysis}\label{section_2}
\subsection{Preliminaries}
Prior to delving into the model, a few explanations regarding the notations used throughout the paper are provided. Lower-case letters, e.g., $x$, denote scalars, lower-case (upper-case) bold letters denote vectors (matrices), e.g., $\mathbf{x}$ ($\mathbf{X}$).
Moreover, $L_1$ ($L_2$) vector norm is defined as $\norm{\mathbf{x}}_1 \doteq \sum_i \left|x_i\right|$ ($\norm{\mathbf{x}}_2 \doteq \sqrt{\sum_{i} x_i^2}$). Similarly, $L_{1,1}\equiv L_{1}$ ($L_{2,2}\equiv L_F$) matrix norm is defined as $\norm{\mathbf{X}}_1\doteq \sum_{i,j} \left|x_{ij}\right|$ ($\norm{\mathbf{X}}_F \doteq \sqrt{\sum_{i,j} x_{ij}^2}$). The nuclear norm of a matrix $\mathbf{X}$, i.e., the sum of its singular values, is defined as $\norm{\mathbf{X}}_{*}$. The Hadamard, i.e., element-wise, product of two matrices $\mathbf{X}$ and $\mathbf{Y}$ is denoted as $\mathbf{X}\odot \mathbf{Y}$. Finally, we provide the following operator definitions which will be utilized in the mathematical derivations required for \modelname.
\begin{itemize}
\item Procrustes operator: $\mathcal{Q}\left(\mathbf{X}\right) \doteq \mathbf{U}\mathbf{V}^T$, where $\mathbf{U}$ and $\mathbf{V}$ are given by the rank-$r$ Singular Value Decomposition (SVD) of $\mathbf{X}$, i.e., $\mathbf{X} = \mathbf{U}\boldsymbol{\Sigma}\mathbf{V}^T$.
\item Shrinkage operator: $\mathcal{S}_{\tau}(\sigma) \doteq \textrm{sgn}\left(\sigma\right)\max \left(\left|\sigma\right|-\tau, 0\right)$
\item Singular Value Thresholding (SVT) operator: $\mathcal{D}_{\tau} \doteq \mathbf{U}\mathcal{S}_{\tau}\mathbf{V}^T$, where $\mathbf{U}$ and $\mathbf{V}$ are given by the rank-$r$ SVD of $\mathbf{X}$, i.e., $\mathbf{X} = \mathbf{U}\boldsymbol{\Sigma}\mathbf{V}^T$.
\end{itemize}

\subsection{Problem formulation}\label{problem_formulation}

Without any loss of generality, suppose that the incomplete, contaminated with gross errors UV maps at-hand are annotated for $J$ attributes (e.g., {\it identity}, {\it age}, etc.), where each attribute may have $M_i,\forall i\in\left\{1,\ldots, J\right\}$, different instantiations (e.g., attribute {\it identity} may have the instantiation {\it Frank Sinatra}, {\it Albert Einstein}, etc.). Moreover, assume that there is a total of $N$ samples in the training set. Aim of \modelname\ is to robustly extract $J$ joint components corresponding to the available attributes during training, an individual component which captures the rest data information that cannot be explained by the $J$ components and a component which captures the gross but sparse errors. Let training data be concatenated in a column-wise manner, i.e., $\mathbf{X}= \begin{bmatrix}\mathbf{x}_1\ldots \mathbf{x}_N\end{bmatrix}$, where $\mathbf{x}_i\in\mathbb{R}^{F\times 1}, i\in\left\{1,\ldots,N\right\}$, is a vectorized form of a facial UV map. Then \modelname\ admits the following decomposition.
\begin{align}
\mathbf{X} = \sum_{i=1}^J \mathbf{S}_i + \mathbf{I} + \mathbf{E},\label{only_components}
\end{align}
where $\mathbf{S}_i, i\in\left\{1,\ldots, J\right\}$, are the $J$ shared components for every attribute, $\mathbf{I}$ is the individual component and $\mathbf{E}$ is the error component. 

Nevertheless, $\mathbf{S}_i, i\in \left\{1,\ldots,J\right\}$, must have a specific low-rank structure which accounts for the different instantiations of each attribute. That is, every attribute should be rendered by a \textsl{base} and subsequently every corresponding instantiation be rendered by a \textsl{selector} on that base. Therefore, \eqref{only_components} is re-formulated as follows.
\begin{align}
\mathbf{X} = \sum_{i=1}^J \mathbf{F}_i\mathbf{H}_i + \mathbf{G} + \mathbf{E},\label{decomposition}
\end{align}
where $\mathbf{F}_i\in\mathbb{R}^{F\times M_i}, i\in\left\{1,\ldots,J\right\}$, are the \textsl{bases} that render each attribute and $\mathbf{H}_i\in\mathbb{R}^{M_i\times N}, i\in\left\{1,\ldots,J\right\}$, are comprised of the \textsl{shared selectors}, i.e., $\mathbf{H}_i \doteq \begin{bmatrix} \mathbf{h}_{i,1} \ldots \mathbf{h}_{i,M_i}\end{bmatrix}$, which render a specific instantiation for an attribute (e.g., assuming that base $\mathbf{F}_i$ renders attribute {\it identity}, then $\mathbf{h}_{M_i}$ would render a particular instantiation of this attribute, e.g., {\it Albert Einstein}). It should be noted that data which bear the same instantiation for a particular attribute (e.g., multiple data with instantiation {\it Albert Einstein}) have the same selector. Furthermore, low-rank base $\mathbf{G}\in\mathbb{R}^{F\times N}$, renders the individual variation for all of the images in the training set that cannot be explained by the existing attributes. Finally, $\mathbf{E}\in\mathbb{R}^{F\times N}$ encapsulates gross errors (such as occlusions, pixel corruptions, etc.) for all of the data samples in the training set.

In order to recover components $\left\{\mathbf{F}_i\mathbf{H}_i\right\}_{i=1}^J$ and $\mathbf{G}$ which are as informative as possible, the error term which accounts for the existence of gross but sparse errors in the visible parts of the UVs has to be minimized.  This is equivalent to minimizing the $L_1$ norm of the error term \cite{donoho2006most} for the visible parts of the UV maps. The problem is then formulated as follows. 
\begin{align}
&\min_{\theta} \norm{\mathbf{W}\odot\mathbf{E}}_1,\nonumber\\
\textrm{s.t.}\ \mathbf{X} &= \sum_{i=1}^J \mathbf{F}_i\mathbf{H}_i + \mathbf{G} + \mathbf{E}\label{minimization}\\
&\left\{\mathbf{F}_i^T\mathbf{F}_i=\mathbf{I}\right\}_{i=1}^J, \rank\left(\mathbf{G}\right) = R,\nonumber
\end{align}
where $\theta = \left\{\mathbf{F}_i,\mathbf{H}_i,\mathbf{G}\right\}, i\in\left\{1,\ldots,J\right\}$ and $R, R<\min(F,N)$, is a hyper-parameter. Moreover, $\mathbf{W}\doteq \begin{bmatrix} \mathbf{w}_1\ldots \mathbf{w}_N\end{bmatrix}$, where $\mathbf{w}_i\in\left\{0,1\right\}^{F\times 1}, i\left\{1,\ldots,N\right\}$, is the corresponding vectorized occlusion mask for each UV map $\mathbf{x}_i$. The visible (missing) pixels for each UV map correspond to ones (zeros) in each matching occlusion mask. Orthonormalization constraints on the bases $\left\{\mathbf{F}_i\right\}_{i=1}^J$ facilitate the recovery of unique and identifiable selectors. Because of the fact that $R$ is a hyper-parameter, it requires a large number of experiments to estimate the optimal rank for $\mathbf{G}$. Since $R$ is upper bounded, the following relaxed decomposition can be used to automatically recover the optimal rank for $\mathbf{G}$.
\begin{align}
&\min_{\theta}\lambda\norm{\mathbf{W}\odot\mathbf{E}}_1 + \norm{\mathbf{G}}_{*},\nonumber\\
\textrm{s.t.}\ \mathbf{X} &=\sum_{i=1}^J\mathbf{F}_i\mathbf{H}_i + \mathbf{G} + \mathbf{E},\label{minimization_extended}\\
&\left\{\mathbf{F}_i^T\mathbf{F}_i=\mathbf{I}\right\}_{i=1}^J,\nonumber
\end{align}
where the nuclear norm of $\mathbf{G}$ is introduced as a convex surrogate of the rank function \cite{candes2011robust} and $\lambda>0$ is a regularizer. 

\subsection{Mathematical derivations}
Because problem \eqref{minimization_extended} is separable, we adopt a alternating optimization scheme to find the updates for every parameter. The corresponding partially Augmented Lagrangian for \eqref{minimization_extended} may then be written as
\begin{align}
\mathcal{L}\left(\theta\right) &= \lambda\norm{\mathbf{W}\odot\mathbf{E}}_1 + \norm{\mathbf{G}}_{*} - \frac{1}{2\mu}\norm{\mathbf{\boldsymbol{\Lambda}}}_F^2 +\nonumber\\ 
&\frac{\mu}{2}\norm{\mathbf{X} - \sum_{i=1}^J\mathbf{F}_i\mathbf{H}_i - \mathbf{G} - \mathbf{E} + \frac{\boldsymbol{\Lambda}}{\mu}}_F^2,\label{augmented_lagrangian}\\
\textrm{s.t.} &\left\{\mathbf{F}^T_i\mathbf{F}_i=\mathbf{I}\right\}_{i=1}^J,\nonumber
\end{align}
where $\theta = \left\{\mathbf{F}_i,\mathbf{H}_i,\mathbf{G},\boldsymbol{\Lambda}\right\}, i\in\left\{1,\ldots, J\right\}$. Problem \eqref{augmented_lagrangian} is minimized by employing the Alternating Direction Method of Multipliers (ADMM) \cite{gabay1976dual, bertsekas2014constrained}. Comprehensive derivations of the optimization problems and the complete algorithm for solving \eqref{minimization_extended} are provided in the supplementary material. The algorithm terminates when the iterations reach a predefined max value or a convergence criterion is met. The convergence criterion is met when the normalized reconstruction error, i.e., $\norm{\mathbf{X} - \sum_{i=1}^J\mathbf{F}_{i}[t]\mathbf{H}_{i}[t] - \mathbf{G}[t] - \mathbf{W}\odot\mathbf{E}[t]}_F/\norm{\mathbf{X}}_F$ is less than a predefined threshold $\epsilon$. The ADMM iteration reads as follows.

\textbf{Update the primal variables}:\\
For obtaining $\mathbf{H}_i, i\in\left\{1,\ldots,J\right\}$, where, as previously mentioned, $\mathbf{H}_i =  \begin{bmatrix}\mathbf{h}_{i,1}\ldots \mathbf{h}_{i,M_i}\end{bmatrix}$, we need to solve individually for every $\mathbf{h}_{i,j}, j\in\left\{1,\ldots,M_i\right\}$. Based on \eqref{augmented_lagrangian}, the solution is given by minimizing
\begin{align}
\mathbf{h}_{i,j}[t+1] &= \argmin_{\mathbf{h}_{i,j}[t]}\Bigg\lVert\mathbf{X}-\sum_{k=1}^{J}\mathbf{F}_{k}[t]\mathbf{H}_{k}[t] - \mathbf{G}[t] -\nonumber\\
&-\mathbf{E}[t] + \frac{\boldsymbol{\Lambda}[t]}{\mu[t]}\Bigg\rVert_F^2.\label{minimization_h}
\end{align}
Problem \eqref{minimization_h} admits a closed-form solution, which is
\begin{align}
\mathbf{h}_{i,j}[t+1] &= \frac{1}{N_{i,j}}\mathbf{F}_{i}^T[t]\Bigg(\mathbf{X}^{i,j}-\sum_{k=1, k\neq i}^J\mathbf{F}_{k}^{i,j}[t]\mathbf{H}_{k}^{i,j}[t] -\nonumber\\
&\mathbf{G}^{i,j}[t] - \mathbf{E}^{i,j}[t] + \frac{\boldsymbol{\Lambda}^{i,j}[t]}{\mu[t]}\Bigg)\cdot\mathbf{1},
\end{align}
where superscript $\left\{i,j\right\}$ means that only the $N_{i,j}$ columns corresponding each time to the $j$-th instantiation of the $i$-th attribute are considered (e.g., columns corresponding to data annotated for attribute {\it identity} and instantiation {\it Albert Einstein}) and $\mathbf{1}$ is a column vector of $N_{i,j}$ ones.

For deriving subspace $\mathbf{F}_i, i\in\left\{1,\ldots,J\right\}$, the following needs to be solved.
\begin{align}
\mathbf{F}_{i}[t+1] = &\argmin_{\mathbf{F}[t]}\Bigg\lVert\mathbf{X}-\sum_{j=1}^J\mathbf{F}_{j}[t]\mathbf{H}_{j}[t+1]-\mathbf{G}[t] -\nonumber\\
&- \mathbf{E}[t] + \frac{\boldsymbol{\Lambda}[t]}{\mu[t]}\Bigg\rVert_F^2,\label{minimization_F}\\
&\textrm{s.t.}\ \left\{\mathbf{F}_i^T\mathbf{F}_i = \mathbf{I}\right\}_{i=1}^{J}.\nonumber
\end{align}
In order to solve \eqref{minimization_F}, we rely on the Procrustes Operator $\mathcal{Q}$ and the Lemma introduced next.\\
\textbf{Lemma}: The constraint minimization problem
\begin{align}
\mathbf{\boldsymbol{\Omega}}^{*} &= \argmin_{\boldsymbol{\Omega}}\norm{\boldsymbol{\Omega}\mathbf{A} - \mathbf{B}}_F^2\label{orthogonal_procrustes}\\
&\textrm{s.t.}\ \boldsymbol{\Omega}^T\boldsymbol{\Omega}=\mathbf{I}\nonumber,
\end{align}
has a closed-form solution \cite{schonemann1966generalized} of the form $\boldsymbol{\Omega}^{*} = \mathcal{Q}\left(\mathbf{B}\mathbf{A}^T\right)$.
As a result, the solution for \eqref{minimization_F}, taking into account \eqref{orthogonal_procrustes}, is
\begin{align}
\mathbf{F}_{i}[t+1] &= \mathcal{Q}\Bigg[\Bigg(\mathbf{X} - \sum_{j=1, j\neq i}^{J}\mathbf{F}_{j}[t]\mathbf{H}_{j}[t+1] - \mathbf{G}[t] -\nonumber\\
&-\mathbf{E}[t] + \frac{\boldsymbol{\Lambda}[t]}{\mu[t]}\Bigg) \cdot\mathbf{H}_{i}[t+1]^T\Bigg].
\end{align}

For obtaining subspace $\mathbf{G}$, the following needs to be solved.
\begin{align}
\mathbf{G}[t+1] &= \argmin_{\mathbf{G}[t]}\Bigg[\frac{\mu[t]}{2}\Bigg\lVert\mathbf{X} - \sum_{i=1}^{J}\mathbf{F}_{i}[t+1]\mathbf{H}_{i}[t+1] -\nonumber\\
&-\mathbf{G}[t] - \mathbf{E}[t] + \frac{\boldsymbol{\Lambda}[t]}{\mu[t]}\Bigg\rVert_F^2 + \norm{\mathbf{G}[t]}_{*}\Bigg].\label{minimization_G}
\end{align}
Problem \eqref{minimization_G} is solved utilizing the SVT operator $\mathcal{D}$. The solution is
\begin{align}
\mathbf{G}[t+1] &= \mathcal{D}_{\frac{1}{\mu[t]}}\Bigg[\mathbf{X} - \sum_{i=1}^J\mathbf{F}_{i}[t+1]\mathbf{H}_{i}[t+1] - \mathbf{E}[t] +\nonumber\\
&+ \frac{\boldsymbol{\Lambda}[t]}{\mu[t]} \Bigg].
\end{align}
For obtaining $\mathbf{E}$, the following problem needs to solved.
\begin{align}
&\mathbf{E}[t+1] = \argmin_{\mathbf{E}[t]}\Bigg[\lambda\norm{\mathbf{W}\odot\mathbf{E}[t]}_1 + \frac{\mu[t]}{2}\Bigg\lVert\mathbf{X} -\nonumber\\
&- \sum_{i=1}^J\mathbf{F}_{i}[t+1]\cdot\mathbf{H}_{i}[t+1] -\mathbf{G}[t+1] -\mathbf{E}[t] + \frac{\boldsymbol{\Lambda}[t]}{\mu[t]}\Bigg\rVert_F^2\Bigg]\label{minimization_E}
\end{align}
Problem \eqref{minimization_E} is solved utilizing the Shrinkage operator $\mathcal{S}$. The solution is
\begin{align}
\mathbf{E}[t+1] &= \mathbf{W}\odot\mathbf{Y} + \overline{\mathbf{W}}\odot\Bigg[\mathbf{X} - \sum_{i=1}^J\mathbf{F}_{i}[t+1]\mathbf{H}_{i}[t+1] -\nonumber\\
&- \mathbf{G}[t+1] + \frac{\boldsymbol{\Lambda}[t]}{\mu[t]}\Bigg],
\end{align}
where
\begin{align}
\mathbf{Y} &= \mathcal{S}_{\frac{\lambda}{\mu[t]}}\Bigg[\mathbf{X} - \sum_{i=1}^J\mathbf{F}_{i}[t+1]\mathbf{H}_{i}[t+1] -\nonumber\\
&- \mathbf{G}[t+1] + \frac{\boldsymbol{\Lambda}[t]}{\mu[t]}\Bigg]
\end{align}
and $\overline{\mathbf{W}}$ is the complement of $\mathbf{W}$.\\ \\
\textbf{Update the Lagrange multiplier and} $\mu_t$ \textbf{parameter}:
\begin{align}
\boldsymbol{\Lambda}[t+1] &= \boldsymbol{\Lambda}[t] + \mu[t]
\Bigg(\mathbf{X} - \sum_{i=1}^J\mathbf{F}_{i}[t+1]\mathbf{H}_{i}[t+1] -\nonumber\\
&- \mathbf{G}[t+1] - \mathbf{E}[t+1]\Bigg),\\
\mu[t+1] &= \min\left(\rho\mu[t], \mu_{\max}\right).
\end{align}

Regarding the theoretical convergence of the ADMM algorithm presented previously, there is no proof when ADMM is utilized in settings with more than two blocks of variables. Nevertheless, ADMM provides good results in non-linear optimization problems \cite{peng2012rasl}. Furthermore, experimental evaluation of MA-RCA in a number of different tasks on ``in-the-wild'' data admits that the derived solutions constitute a good approximation. 

\subsection{Reconstruction of a test image}\label{reconstruction}
After the bases and the selectors have been recovered as described in Section \ref{problem_formulation}, they may be utilized in order to recover the shared and individual components of a test image. Then the said components can be utilized in experiments such as completion of missing UV parts and {\it joint} transfer of a facial test image to another {\it age}, {\it identity} or {\it illumination}, as demonstrated in Section \ref{experiments}.

Without any loss of generality, assume a test UV map $\mathbf{y}$, which may be decomposed in the shared and individual components as follows.
\begin{align}
\mathbf{y} = \sum_{i=1}^J \mathbf{F}_i\hat{\mathbf{h}}_i + \mathbf{K}\hat{\mathbf{w}} + \hat{\mathbf{\epsilon}},\label{recon_decomposition}
\end{align}
where $\mathbf{K}$ is the linear span of $\mathbf{G}$, given by applying the rank-$r$ SVD on $\mathbf{G}$. In the most general case, optimal selectors $\left\{\hat{\mathbf{h}}_i\right\}=_{i=1}^J$ and $\hat{\mathbf{w}}$ must be extracted by minimizing the sparse error term $\hat{\boldsymbol{\epsilon}}$ corresponding to the visible part of $\mathbf{y}$, for already recovered $\left\{\mathbf{F}_i\right\}_{i=1}^J$ and $\mathbf{G}$. That is, the following needs to be solved. 
\begin{align}
&\min_{\theta} \norm{\mathbf{w}_{y}\odot\hat{\boldsymbol{\epsilon}}}_1\nonumber\\
\textrm{s.t.}\quad \mathbf{y} &= \sum_{i=1}^J \mathbf{F}_i\hat{\mathbf{h}}_i + \mathbf{K}\hat{\mathbf{w}} + \hat{\boldsymbol{\epsilon}},\label{recon_minimization}\\
\hat{\mathbf{y}} &= \sum_{i=1}^J \mathbf{F}_i\hat{\mathbf{h}}_i + \mathbf{K}\hat{\mathbf{w}},\nonumber
\end{align}
where $\mathbf{w}_y$ is the occlusion mask corresponding to the test UV map $\mathbf{y}$ and $\hat{\mathbf{y}}$ is the reconstructed facial UV map.
In the case where, e.g., transfer of a test image to a specific age is required, the selector corresponding to the specific age will be fixed (i.e., the corresponding selector found during the training process in Section \ref{problem_formulation} is utilized). Problem \eqref{recon_minimization} is solved by employing the ADMM. The algorithm and complete derivations for solving problem \eqref{recon_minimization} are provided in the supplementary material. 

\section{Experiments}\label{experiments}

The experimental evaluation of MA-RCA against other state-of-the-art algorithms is carried out via a series of experiments such as: a) completion of UV maps with missing values on data captured in ``in-the-wild'' conditions, b) age-progression on data captured in controlled as well as ``in-the-wild'' conditions, c) joint {\it illumination} and {\it identity} transfer on data captured in controlled as well as ``in-the-wild'' conditions. Moreover, we demonstrate how MA-RCA can be employed to produce data with specific characteristics (e.g., ``in-the-wild'' images illuminated from various angles) that can be then utilized to train deep networks for tailored applications such as illumination transfer. 

In order to extract the incomplete facial UV maps that have been used in our algorithm we have fitted the various databases with a 3DMM. The 3DMM fitting process that has been used was the one in \cite{booth20173d}, which is publicly available. The shape and camera parameters were used in order to sample the texture and compute the occlusion masks in the fitted image.

For the experimental evaluations, databases Multi-PIE \cite{gross2010multi} and AgeDB \cite{moschoglou2017agedb} were utilized to train MA-RCA. Multi-PIE is a database captured under controlled lab conditions and thus the images do not contain gross-errors attributed to e.g., occlusions. Nevertheless, Multi-PIE is a multi-attribute database, since it contains labels for attributes such as {\it identity} and {\it illumination}. That renders it suitable to be utilized in MA-RCA to extract bases with respect to e.g., {\it illumination} that can be then used to reconstruct ``in-the-wild'' images with various illumination settings (Section \ref{illumination}). In particular, in the training phase of MA-RCA on Multi-PIE, all of the identities and illuminations available in the database were used. 

AgeDB contains images captured under ``in-the-wild'' conditions (i.e., occlusions, various poses, pixel corruptions are present in the images). Moreover, it is annotated for multiple attributes (i.e., {\it identity}, {\it age}) and thus it is suitable for evaluating MA-RCA. AgeDB was split in six distinct age-groups, namely {\it 21-30}, {\it 31-40}, {\it 41-50}, {\it 51-60}, {\it 61-70} and {\it 71-100}. Then the UVs belonging to each age-group were further split according to attribute {\it identity}. In the training phase of MA-RCA, $90\%$ of the total UVs were kept to extract the bases with respect to attributes {\it identity} and {\it age-groups} and the rest were used for testing.

\subsection{Completion of UV maps with missing values}
Completion of UV maps with missing values is a very challenging task which, to the best of our knowledge, has not been addressed in the literature. MA-RCA is the first technique that can be utilized to handle UV maps with missing values.  In this experiment, AgeDB ``in-the-wild'' \cite{moschoglou2017agedb} was utilized. 

\begin{figure}[t]
\begin{center}
\includegraphics[width=1\linewidth]{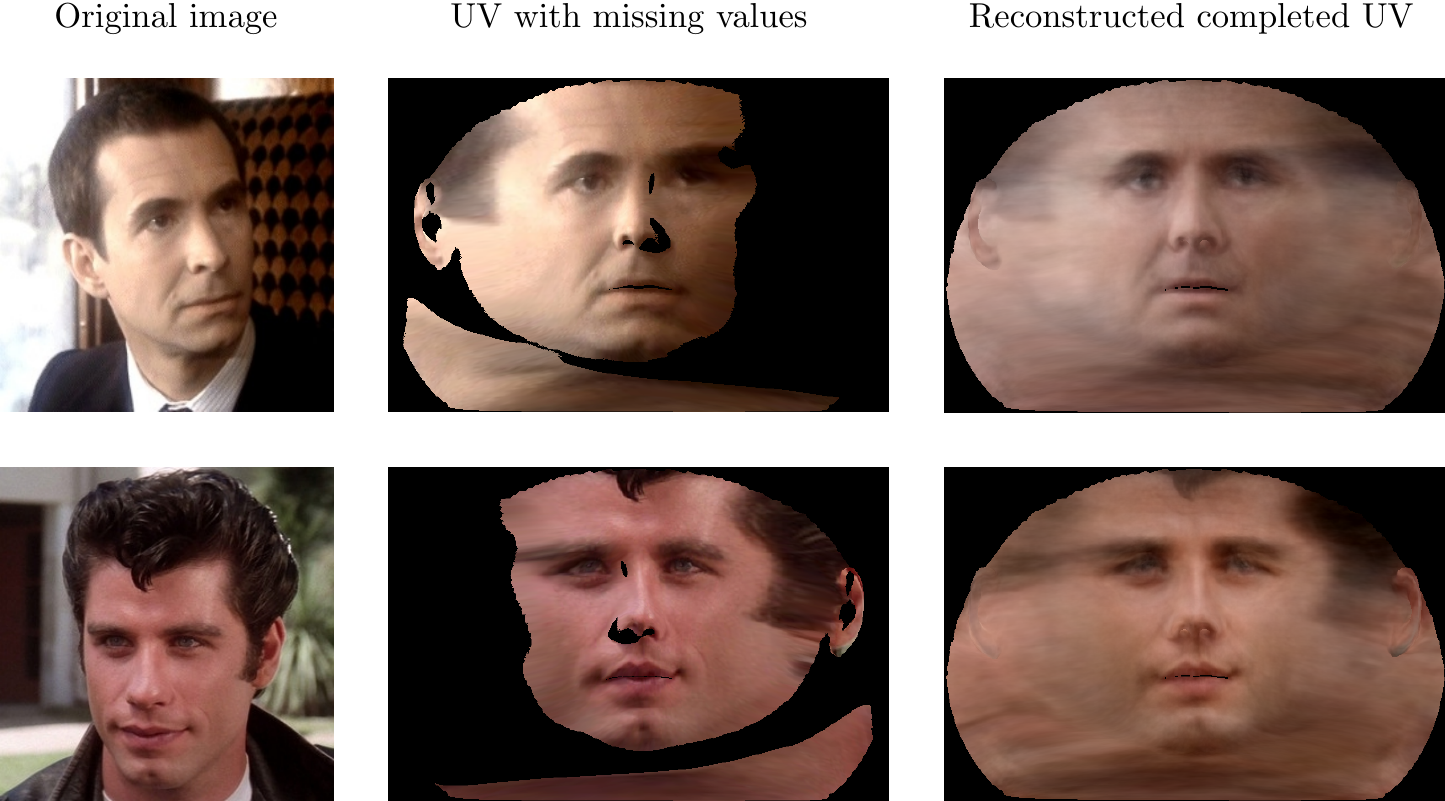}
\end{center}
   \caption{Examples of UV map completion utilizing MA-RCA. A 3DMM \cite{booth20173d} is fitted in the original image and the UV map with missing values and sparse errors is extracted. Finally, the complete UV map is reconstructed utilizing MA-RCA.}
\label{fig:3}
\end{figure}

\begin{figure}[t]
\begin{center}
\includegraphics[width=1\linewidth]{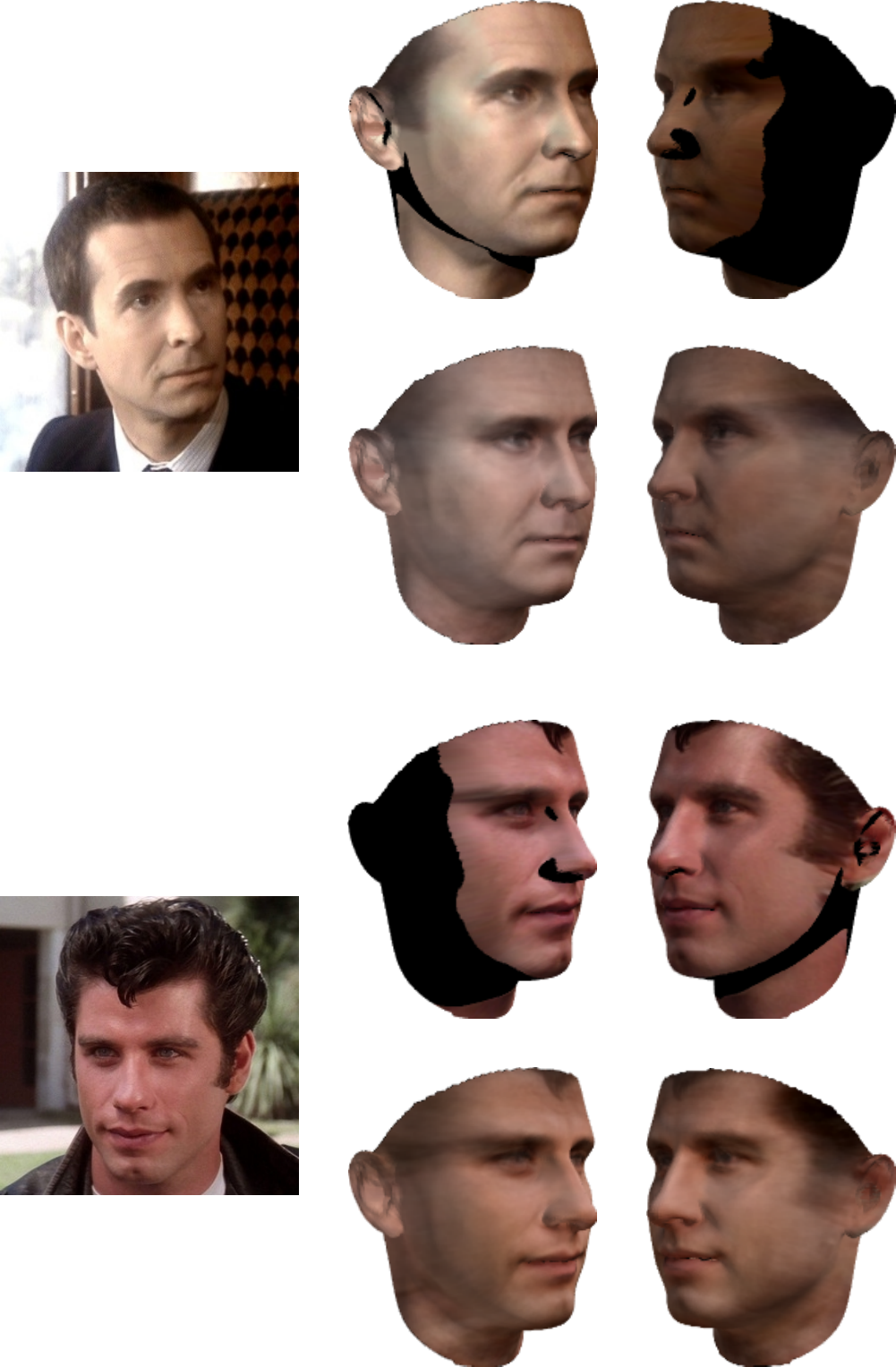}
\end{center}
   \caption{Examples of 3D texture completion utilizing MA-RCA. Multiple views are visualized to demonstrate the texture completion under various angles.}
\label{fig:4}
\end{figure}

For the testing phase, a random, incomplete, contaminated with gross but sparse errors UV map which did not belong to the training set was chosen and reconstructed following the process described in Section \ref{reconstruction}. As it is evident in Fig. \ref{fig:3} as well as Fig. \ref{fig:4}, MA-RCA successfully fills the missing, occluded parts in the original UV map as well as the missing parts in the corresponding 3D facial textures.

\subsection{Age-progression ``'in-the-wild''}
Age-progression ``in-the-wild'' entails the task of rendering a facial image of a subject at various ages. It is arguably a very challenging task in Computer Vision, since ``in-the-wild'' images have been captured in uncontrolled conditions (e.g., different illuminations and poses, self-occlusions, etc.). AgeDB \cite{moschoglou2017agedb} was utilized in this experiment, since it is a manually collected ``in-the-wild'' age database with accurate {\it age} and {\it identity} labels and hence the extracted age-groups and identity bases will contain no errors due to incorrect annotations. 

In the testing phase, a random UV map which did not belong to the training set was chosen and reconstructed for various ages following the process described in Section \ref{reconstruction}.

\begin{figure}[t]
\begin{center}
\includegraphics[width=1\linewidth]{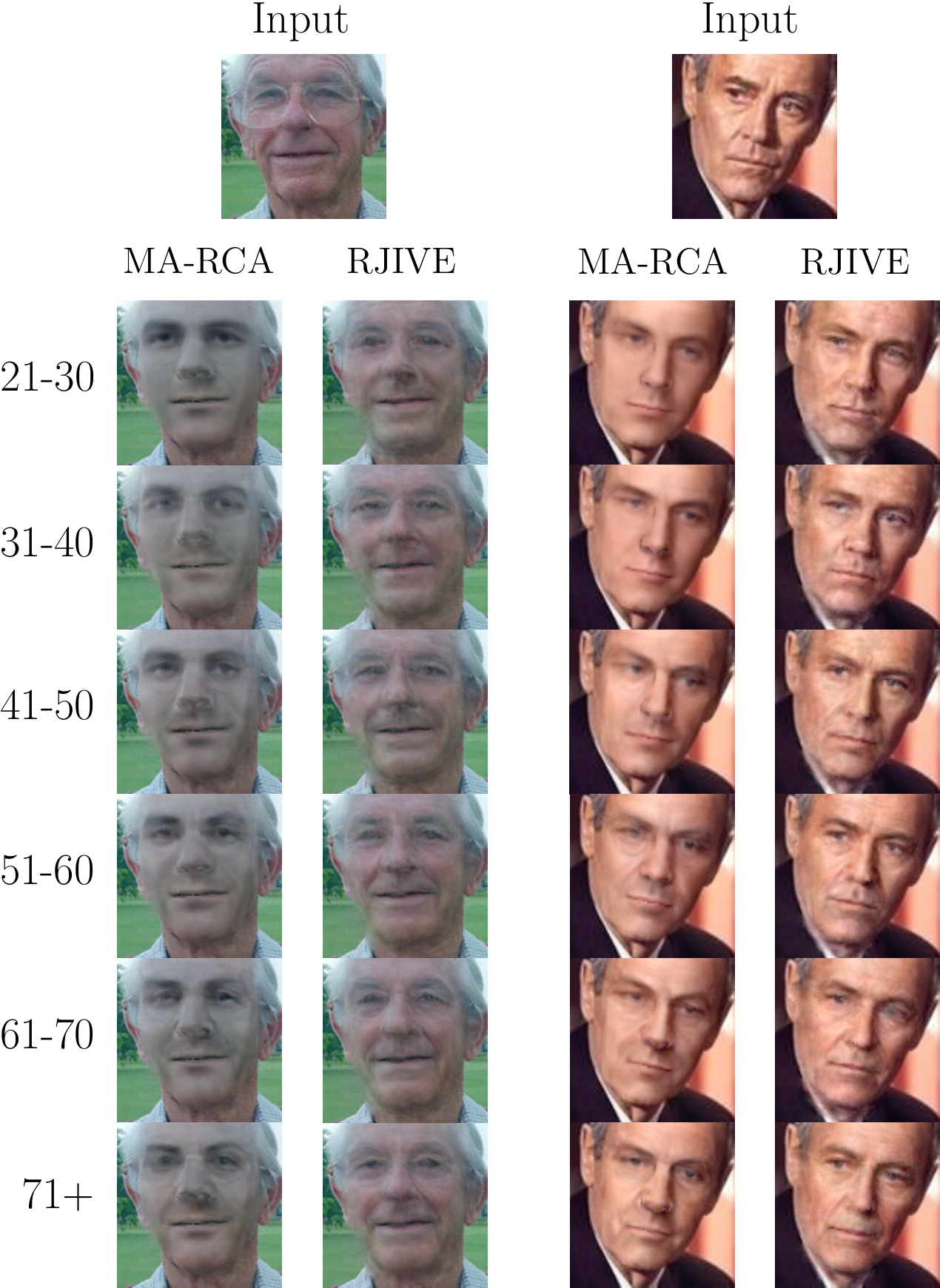}
\end{center}
   \caption{Comparing MA-RCA against state-of-the-art RJIVE in age-progression ``in-the-wild'' experiments. As can be seen in both cases, MA-RCA reconstructions are more realistic compared to the reconstructions utilizing RJIVE. More examples can be found in the supplementary material.}
\label{fig:5}
\end{figure}

\begin{figure}[t]
\begin{center}
\includegraphics[width=1\linewidth]{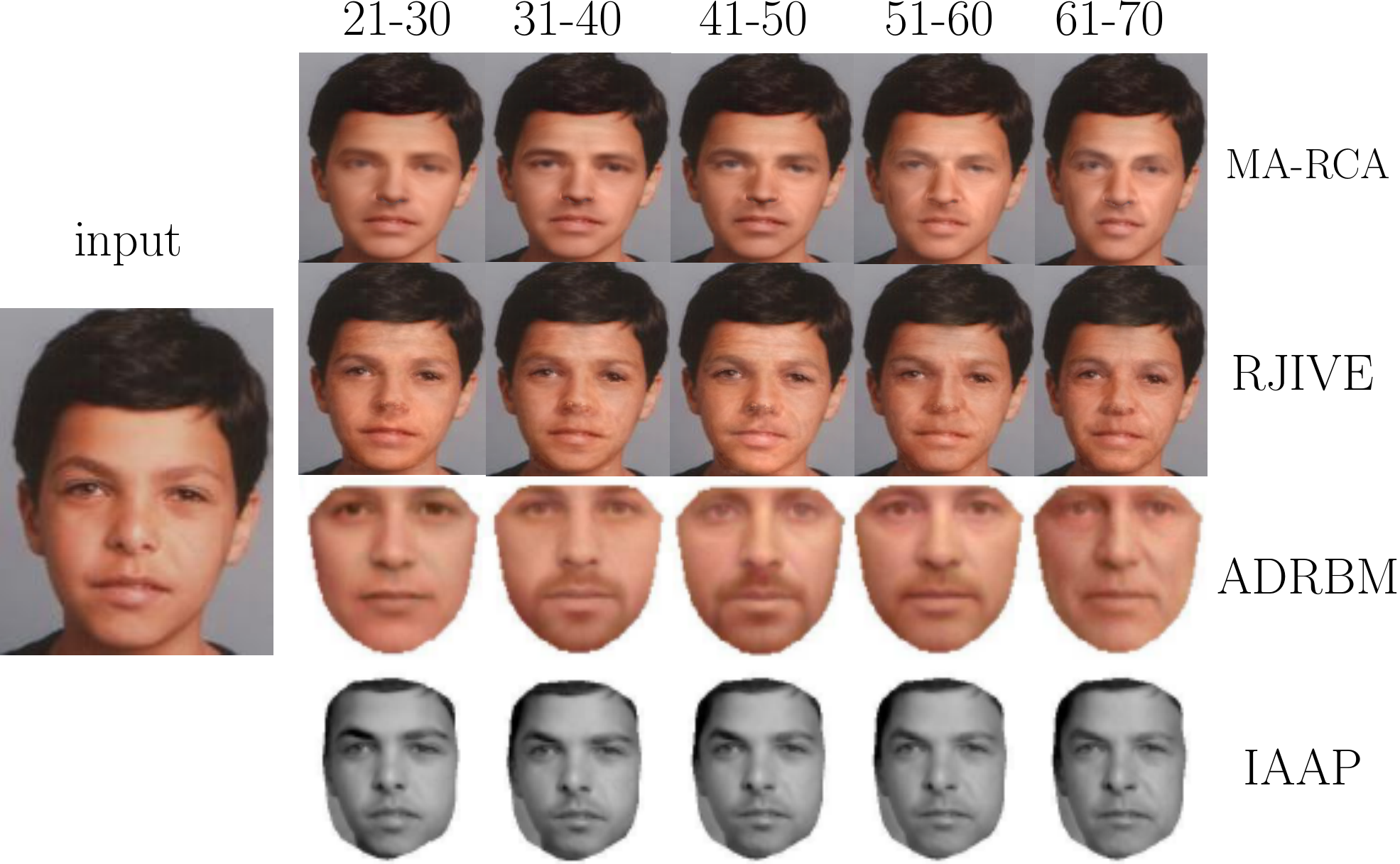}
\end{center}
   \caption{Age-progression results on FG-Net database \cite{cootes2008fg} produced by IAAP, ADRBM, RJIVE and the proposed MA-RCA method. RJIVE and MA-RCA are trained on AgeDB \cite{moschoglou2017agedb}.}
\label{fig:6}
\end{figure}

\begin{figure}[t]
\begin{center}
\includegraphics[width=1\linewidth]{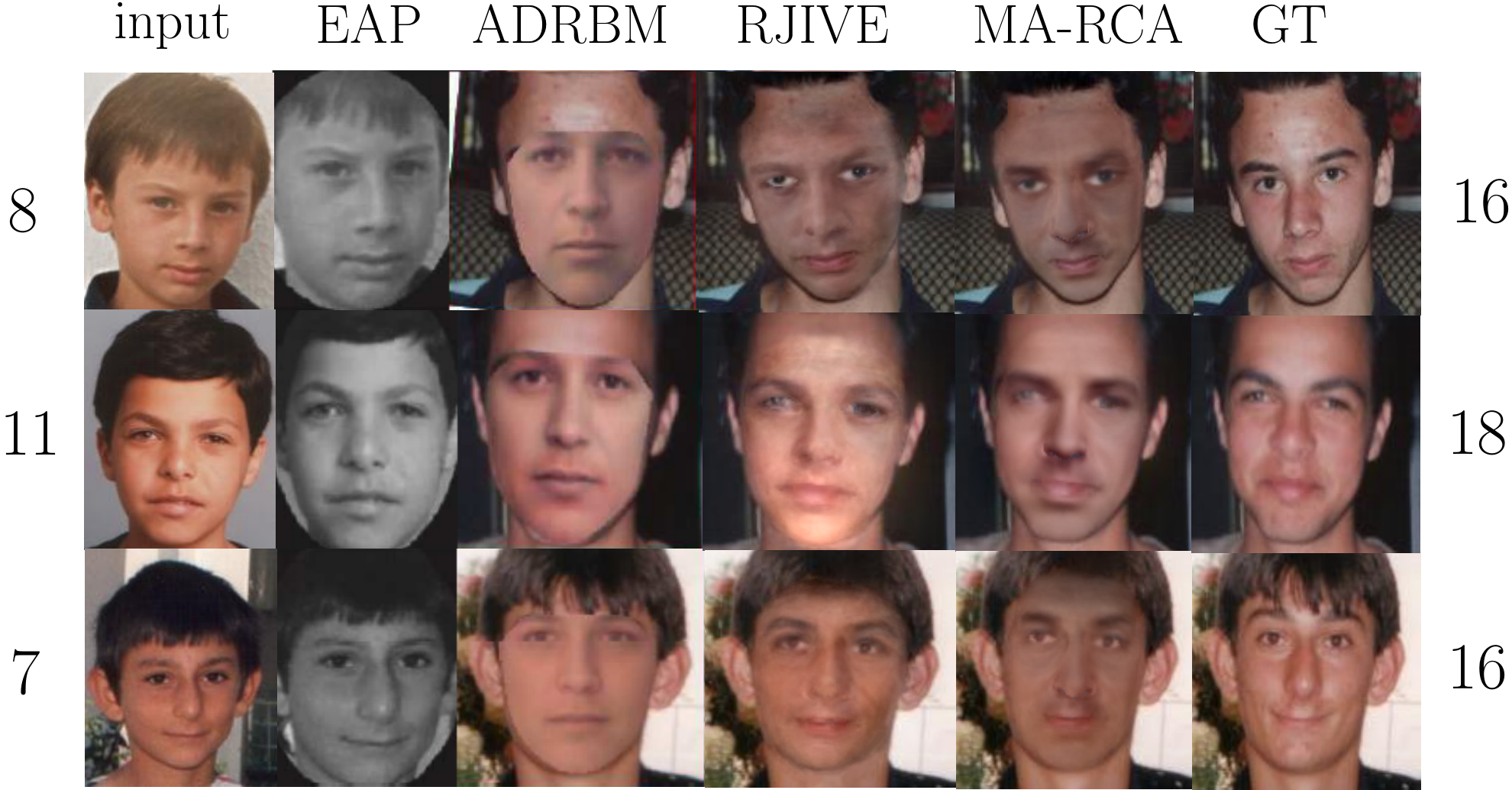}
\end{center}
   \caption{Age-progression results produced by EAP, ADRBM, RJIVE and the proposed MA-RCA method on images of the FG-Net database \cite{cootes2008fg}. MA-RCA and RJIVE are trained on AgeDB \cite{moschoglou2017agedb}.}
\label{fig:8}
\end{figure}

Comparisons against other broadly used age-progression methods are provided in Fig. \ref{fig:6} and Fig. \ref{fig:8}. More specifically, we compare MA-RCA against Illumination Aware Age Progression (IAAP) \cite{kemelmacher2014illumination}, Aging with Deep Restricted Boltzmann Machines (ADRBM) \cite{nhan2016longitudinal}, Exemplar-based Age Progression (EAP) \cite{shen2011exemplar} and RJIVE \cite{sagonas2017robust} . Finally, MA-RCA is compared against the state-of-the-art RJIVE \cite{sagonas2017robust} in Fig. \ref{fig:5}.

\subsection{Multi-attribute transfer ``in-the-wild''}\label{illumination}

In this section, we present a series of experiments under the multi-attribute scenario, i.e., when a test image is reconstructed with more than one attribute transferred at the same time. MA-RCA is the first, to the best of our knowledge, method that can successfully carry out such a task. For this series of experiments, both Multi-PIE \cite{gross2010multi} and AgeDB \cite{moschoglou2017agedb} are utilized. 

\begin{figure}[t]
\begin{center}
\includegraphics[width=1\linewidth]{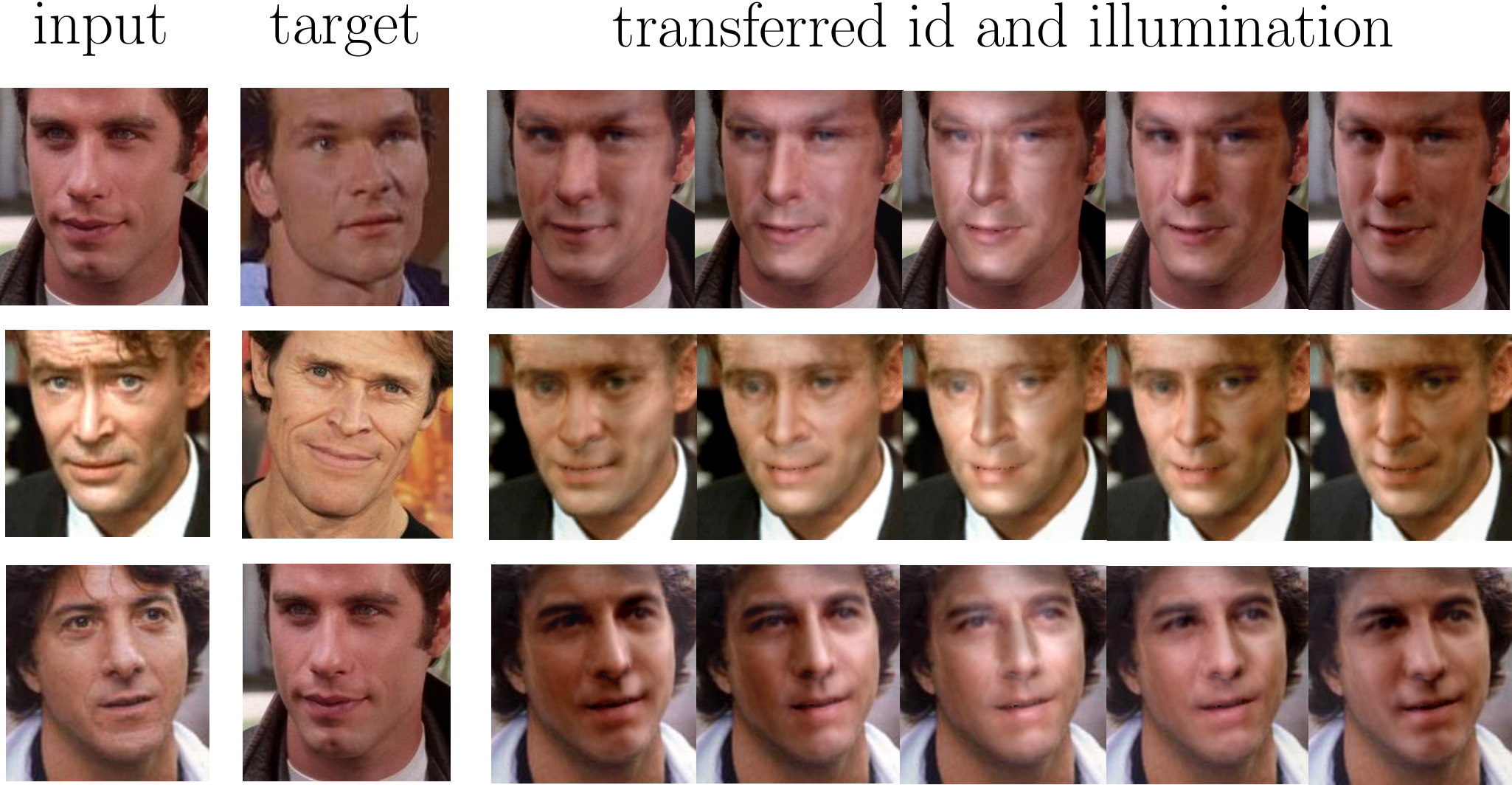}
\end{center}
   \caption{Join transfer of attributes {\it identity} and {\it illumination} ``in-the-wild''. {\it Illumination} base is learned on Multi-PIE \cite{gross2010multi} while {\it identity} base on AgeDB \cite{moschoglou2017agedb}. Final reconstruction of the test image is carried out utilizing both bases.}
\label{fig:10}
\end{figure}

In particular, the {\it illumination} base is extracted from Multi-PIE while the {\it identity} and {\it individual} bases are extracted from AgeDB. During the reconstruction of a test image (Section \ref{reconstruction}) the bases from Multi-PIE as well as AgeDB are utilized. In Fig. \ref{fig:10}, we present how MA-RCA can be utilized to transfer the identity of a particular subject into another one and also transfer the illumination setting at the same time.

\subsection{Utilizing MA-RCA in deep learning applications}
\begin{figure}[t]
\begin{center}
\includegraphics[width=1\linewidth]{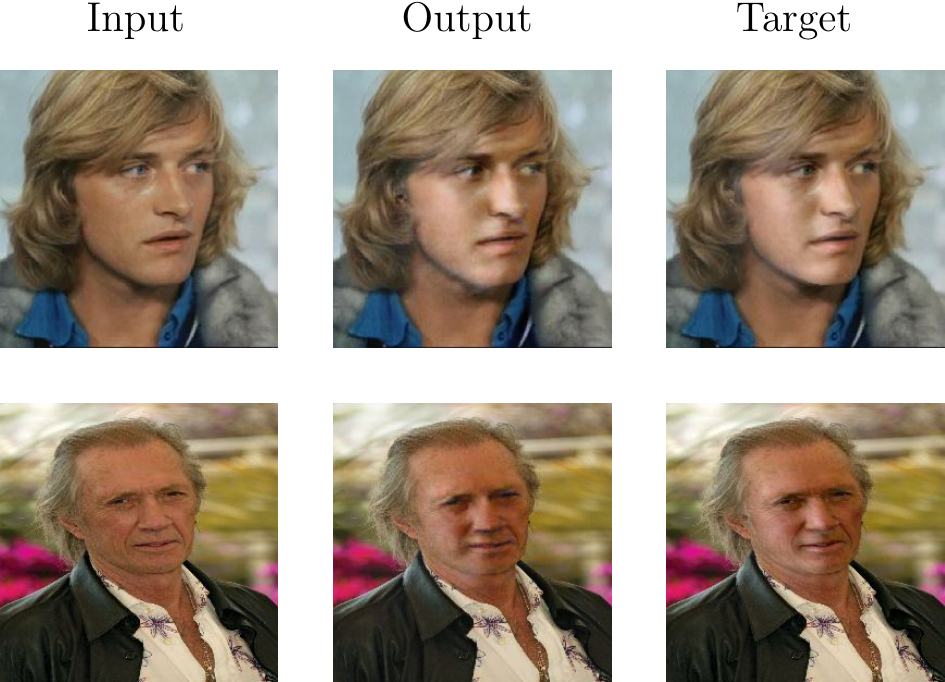}
\end{center}
   \caption{Outputs of pix2pix GAN \cite{isola2016image} in test images (Input). Targets are the illuminated reconstructions of the inputs when MA-RCA is utilized.}
\label{fig:7}
\end{figure}

MA-RCA applications may be also deemed beneficial in the training phase of deep networks. For example, MA-RCA can be utilized to reconstruct ``in-the-wild'' images with a specific illumination setting. The reconstructed illuminated images can then be used in pairs with the corresponding original ``in-the-wild'' data to train a deep generative network (e.g., pix2pix GAN \cite{isola2016image}) for illumination transfer ``in-the-wild''. After the training phase is complete, the trained network can generate illuminated images on test images, similar to the ones MA-RCA would have reconstructed (Fig. \ref{fig:7}). Furthermore, MA-RCA may be also utilized to augment facial datasets (by transferring illumination, age, etc. in images) and thus provide state-of-the-art neural networks with more data during training.

\section{Conclusions}

With the use of 3D face fitting methods we can generate incomplete facial UV maps of the facial textures. The use of incomplete facial UV maps contaminated with gross errors introduces many opportunities and challenges. In particular, since facial UV map lies in a pose free space, linear component analysis techniques can be applied to learn statistical components for various tasks. In this paper, we propose a novel statistical component analysis technique that can tackle the above challenges and at the same time exploit multiple labels of the data at-hand during training. We demonstrate the usefulness of the proposed robust component analysis technique in various tasks including UV map completion on ``in-the-wild'' data, {\it illumination} and {\it identity} transfer, as well as aging.  

\section{Acknowledgements}
S. Moschoglou was supported by the EPSRC
DTA studentship from Imperial College London. E. Ververas was supported by the teaching scholarship from Imperial College London. S. Zafeiriou was partially funded by the EPSRC Project EP/N007743/1 (FACER2VM).

{\small
\bibliographystyle{ieee}
\bibliography{egbib}
}
\end{document}